\documentclass{article}
\usepackage{spconf,amsmath,graphicx,booktabs,array,multirow,siunitx,amssymb,subcaption,float,afterpage, longtable}
\usepackage{adjustbox}
\usepackage{caption}
\usepackage{supertabular}
\title{Evaluating Explainable AI Methods in Deep Learning Models for Early Detection of Cerebral Palsy}
\name{Kimji N. Pellano$^{1}$, Inga Strümke$^{2}$, Daniel Groos$^{1}$, Lars Adde$^{3}$, Espen Alexander F. Ihlen$^{1}$} 
\address{
\begin{tabular}{@{}l@{}}
\small $^{1}$ Department of Neuromedicine and Movement Science, Faculty of Medicine and Health Sciences, \\
\small Norwegian University of Science and Technology, 7034 Trondheim, Norway\\
\small$^{2}$ Department of Computer Science, Faculty of Information Technology and Electrical Engineering, \\
\small Norwegian University of Science and Technology, 7034 Trondheim, Norway\\
\small$^{3}$ Department of Clinical and Molecular Medicine, Faculty of Medicine and Health Sciences,\\
\small Norwegian University of Science and Technology, 7034 Trondheim, Norway\\
\end{tabular}
}

\begin{document}
%\ninept
%
\maketitle
\begin{abstract}
Early detection of Cerebral Palsy (CP) is crucial for effective intervention and monitoring. This paper tests the reliability and applicability of Explainable AI (XAI) methods using a deep learning method that predicts CP by analyzing skeletal data extracted from video recordings of infant movements. Specifically, we use XAI evaluation metrics — namely faithfulness and stability — to quantitatively assess the reliability of Class Activation Mapping (CAM) and Gradient-weighted Class Activation Mapping (Grad-CAM) in this specific medical application. We utilize a unique dataset of infant movements and apply skeleton data perturbations without distorting the original dynamics of the infant movements. Our CP prediction model utilizes an ensemble approach, so we evaluate the XAI metrics performances for both the overall ensemble and the individual models. Our findings indicate that both XAI methods effectively identify key body points influencing CP predictions and that the explanations are robust against minor data perturbations. Grad-CAM significantly outperforms CAM in the RISv metric, which measures stability in terms of velocity. In contrast, CAM performs better in the RISb metric, which relates to bone stability, and the RRS metric, which assesses internal representation robustness. Individual models within the ensemble show varied results, and neither CAM nor Grad-CAM consistently outperform the other, with the ensemble approach providing a representation of outcomes from its constituent models. Both CAM and Grad-CAM also perform significantly better than random attribution, supporting the robustness of these XAI methods. Our work demonstrates that XAI methods can offer reliable and stable explanations for CP prediction models. Future studies should further investigate how the explanations can enhance our understanding of specific movement patterns characterizing healthy and pathological development.
\end{abstract}
\section{Introduction}
\label{sec:introduction}
Cerebral Palsy (CP) is the most common motor disability in children, and it is essential to detect it early for effective early intervention and surveillance \cite{novak2017early}. Machine learning based technologies are increasingly being explored in the early detection of CP due to its potential for more accurate, accessible, and timely diagnoses. Specifically, deep learning methods have shown great potential in medical diagnostics due to their ability to detect complex patterns in large sets of data. For instance, McCay et al.\ developed a deep learning framework that classifies infant movements from RGB videos using extracted pose-based features to identify Fidgety Movements (FMs) \cite{mccay2020abnormal}. Similarly, Groos et al.\ introduced a method leveraging deep learning to predict CP from skeletal data captured in spontaneous infant movements, validated across a multi-center cohort \cite{groos2022development}. Additionally, Zhang et al.\ designed CP-AGCN, a graph convolutional network (GCN) that uses skeletal data from RGB videos and a frequency-binning module to classify CP risks in infants \cite{zhang2022cp}. Gao et al. implemented a deep learning model to automate early CP detection by analyzing FMs in video sequences \cite{gao2023automating}. There are several data modalities that can be analyzed for early CP prediction such as in the sensor fusion approach proposed by Kulvicius et al.~\cite{kulvicius2024deep}, but this paper focuses on analyzing skeletal data extracted from video recordings via pose estimation, noting its broader applicability to areas such as abnormal gait detection \cite{nguyen2016skeleton}, Parkinson's disease gait assessment \cite{ajay2018pervasive}, fall detection \cite{chen2020fall}, and other health-related applications.

Despite its promising potential, the use of AI in medical diagnostics introduces new challenges, including the widely discussed problem of explainability and transparency. Deep learning models' inherent lack of interpretability -- commonly referred to as the `black box' challenge -- is problematic in a medical setting, where clear explanations for diagnoses is a requirement. To build trust in AI-driven diagnostic tools among clinicians and patients, and facilitate possible implementation in clinical use, it is crucial to understand the predictions made by these tools. This need for transparency and trust aligns with the requirements of the EU's recently implemented AI regulation, the AI Act, which classifies applications that could affect the life, safety, and health of people as high-risk \cite{EUAIAct2024}, and thereby requires these applications to provide explanations before they are allowed for deployment.

Prechtl's General Movements Assessment (GMA) is a highly reliable diagnostic tool for early detection of CP and is based upon medical experts observing normal versus abnormal movement patterns \cite{ricci2018feasibility}, yet it faces several challenges. These include the requirement for training the clinicians performing GMA to achieve proficiency in assessment techniques, the subjective nature of visual analysis which can lead to variability in interpretations, the time-intensive process of manually analyzing movements, and long-term costs associated with training and qualifying medical experts \cite{silva2021future}. Using AI can potentially address these issues by automating the detection process of abnormal movement patterns, thereby reducing the burden on the limited number of trained GMA experts, while providing objective and consistent analysis, and significantly reducing the time needed for assessments. However, it is crucial that this does not compromise the interpretability that clinicians value in Prechtl's GMA, ensuring that the insights offered by AI systems are sufficiently explained and thus complementary and understandable from a clinician's perspective.

To aid healthcare providers in understanding AI diagnoses, Explainable AI (XAI) methods can be used. For instance, in medical imaging, XAI has facilitated cancer detection \cite{gulum2021review}. In the CP prediction space, Sakkos et al. introduced a deep learning framework using RGB videos, with visualization showing segmented body parts with movement abnormalities and their contribution to the classification result \cite{sakkos2021identification}. Reflecting on the boom of automated solutions for Prechtl's GMA and the growing trend of incorporating XAI in diverse medical applications, and in anticipation of the legal requirements when deploying AI-assisted medical diagnostic tools, our study aims to bridge a crucial gap: despite the evident progress, the application of XAI in skeleton-based CP diagnosis remains understudied. To assess the trustworthiness of the explanations generated by these methods, we advocate for the use of metrics that objectively evaluate the reliability of the explanations. This is crucial because AI-based analysis of skeletal data could extend to several other previously mentioned high-stakes applications, beyond CP prediction. However, the application of domain-specific knowledge across such a broad spectrum of potential uses could become cumbersome. Therefore, we also suggest adopting the metrics evaluated in this study as foundational benchmarks for assessing XAI techniques, and then supplementing them with specialized domain knowledge to enhance and confirm their validity further.

This study explores the applicability of Class Activation Mapping (CAM) and Gradient-Weighted Class Activation Mapping (Grad-CAM), which are widely used XAI methods in Convolutional Neural Network (CNN) models, to Graph Convolutional Network (GCN)-based models for Cerebral Palsy (CP) prediction. Specifically, we investigate whether these XAI methods can effectively differentiate between important and unimportant body points influencing CP predictions. Additionally, we assess the stability of explanations when the input data undergo minor perturbations. 

The contributions of our study are:

\begin{itemize}
    \item An objective evaluation framework to assess the reliability of different XAI methods used in a specific medical application, which is skeleton-based early detection of CP from infants' spontaneous movements.
    \item The comparative analysis of different XAI methods (CAM and Grad-CAM), providing insight into their effectiveness in the context of early CP detection, which can potentially be used in other high-stakes applications.
    \item Showing the possible use of XAI methods to guide further research in early CP diagnosis. For example, it can potentially be used to discover specific infant movement patterns that are correlated with later CP status that might be complementary information for GMA experts focusing on gestalt infant movements. The evaluation of these XAI methods for this specific application is the first step towards determining the most reliable explanations that can provide valuable insights into specific infant movements.
    \item The application of XAI evaluation metrics to an ensemble of models for CP prediction, providing insight into the collective robustness of the aggregated ensemble and individual model explanations against minor perturbations.
\end{itemize}

\subsection{XAI Methods}
The following section discusses the various XAI methods implemented for the skeleton-based CP prediction model. These methods will subsequently be assessed using the XAI evaluation metrics.

\subsubsection{Class Activation Mapping (CAM)}

CAM was originally introduced as a method for identifying important pixels in an image \cite{zhou2016learning} as determined by a CNN model. It projects the model's output layer weights onto a convolutional layer's feature maps (usually from the final layer), creating a heatmap that highlights the areas influencing the network's predictions,

 \begin{equation}
\text{e}_{\text{X, CAM}} = \sum_n w_n^{class} F^n \,.
\label{eq:cam}
\end{equation}

CAM can be generalized and applied to other convolution-based models if the weights after Global Average Pooling (GAP) for a specific $class$ and the nth feature map $F^n$ are multiplied, as shown in Equation~\eqref{eq:cam}. For example, in human activity recognition (HAR) using 3D skeleton graphs as input to a GCN, important body points in the skeleton data can be highlighted, as shown by Song et al. with their work on EfficientGCN \cite{song2022constructing}.

\subsubsection{Gradient-Weighted Class Activation Mapping (Grad-CAM) }
CAM has several weaknesses, such as the lack of flexibility in model architecture due to the need for a GAP layer and a Fully Connected (FC) layer for classification. Gradient-weighted Class Activation Mapping (Grad-CAM) \cite{selvaraju2017grad} addresses this by using gradients entering the FC layer instead of weights, calculated via
\begin{equation}
\text{e}_{\text{X, Grad-CAM}} = \sum_n \alpha_n^{class} F^n \,,
\label{eq:gradcam}
\end{equation}
thus making it adaptable to various CNN architectures. Since the introduction of Grad-CAM, many other CAM-based methods have been introduced to further improve the original CAM. Similarly to CAM, Grad-CAM has been shown to be applicable to GCN-based HAR using skeleton data, as illustrated by Das et al. \cite{das2022gradient}.

\subsection{XAI Evaluation Metrics}
In our previous work \cite{pellano2024movements}, we explored various metrics proposed in \cite{agarwal2022openxai} to evaluate explanations from different XAI methods within the context of skeleton-based HAR. Building on these foundations, this paper aims to extend these evaluation techniques to a specific medical application, which is skeleton-based CP prediction. The following subsections give a brief overview of the applied metrics, illustrating their applicability and relevance in a specific medical use-case. Consider \( X \) as the original input data, with its corresponding explanation \( e_X \), and let \( f(\cdot) \) denote the model's output, representing CP risk. Then, \( X' \) is the perturbed version of \( X \), and \( e'_X \) stands for its corresponding explanation following the perturbation. Top-$k$ refers to the most important features in the input data, while non-top-$k$ refers to the remaining, less important features.

\subsubsection{Faithfulness}
Fidelity or faithfulness \cite{alvarez2018towards,zhou2021evaluating,markus2021role} checks whether an explanation accurately identifies the features that influence a model's predictions. The Prediction Gap on Important features (PGI) in Equation~\eqref{eq:pgi} quantifies the prediction change when key features are altered, while the Prediction Gap on Unimportant features (PGU) in Equation~\eqref{eq:pgu} measures the change when minor features are modified. Ideally, when important features are perturbed, there should be a significant change in the model prediction. Conversely, perturbing unimportant features should have little effect on the prediction. By this logic, a good XAI method should identify both important and unimportant features in a way that results in a high PGI and a low PGU.

\begin{equation}
PGI(X, f, e_X, k) = \mathbb{E}_{X' \sim \text{perturb}(X, e_X, \text{top-}k)}[|f(X) - f(X')|]
\label{eq:pgi}
\end{equation}
\begin{equation}
PGU(X, f, e_X, k) = E_{X' \sim \text{perturb}(X, e_X, \text{non top-}k)}[|f(X) - f(X')|]
\label{eq:pgu}
\end{equation}

\subsubsection{Stability}

Stability or robustness \cite{alvarez2018robustness,agarwal2022rethinking}, measures how much an explanation changes relative to changes in the model's input, output, or internal representation when the original input data undergo minor perturbations, as shown in Equations~\eqref{eq:ris}--\eqref{eq:rrs}. Relative Input Stability (RIS)  includes three components for each of the model's multiple input streams: position, velocity, and bone, referred to as RISp, RISv, and RISb, respectively. ROS refers to Relative Output Stability and RRS refers to Relative Representation Stability. To better understand these concepts in relation to the variables in the model, please refer to Fig.~\ref{fig:flow}, which illustrates the main architecture of the network. \(\mathcal{L}_X\) in Equation~\eqref{eq:rrs} denotes the model's internal representation, which in our study is the logits from the layer preceding the softmax activation function. A stability score of zero is ideal, as it means that visually imperceptible perturbations do not change the explanation at all. 

\begin{equation}
\begin{split}
\operatorname{RIS}\left(X, X^{\prime}, e_{X}, e_{X^{\prime}}\right) &= \max_{X^{\prime}} \frac{\left\|\frac{\left(e_{X}-e_{X^{\prime}}\right)}{e_{X}}\right\|_p}{\max \left(\left\|\frac{\left(X-X^{\prime}\right)}{X}\right\|_p, \epsilon_{\min }\right)} \,, \\
&\quad \forall X^{\prime} \text{ s.t. } X^{\prime} \in \mathcal{N}_{X} ; f(X) = f(X^{\prime})
\end{split}
\label{eq:ris}
\end{equation}

\begin{equation}
\begin{split}
\operatorname{ROS}\left(X, X^{\prime}, e_{X}, e_{X^{\prime}}\right) &= \max_{X^{\prime}} \frac{\left\|\frac{\left(e_{X}-e_{X^{\prime}}\right)}{e_{X}}\right\|_p}{\max \left(\left\|\frac{\left(f(X)-f\left(X^{\prime}\right)\right)}{f(X)}\right\|_p, \epsilon_{\min }\right)} \,,\\
&\quad \forall X^{\prime} \text{ s.t. } X^{\prime} \in \mathcal{N}_{X} ; f(X) = f(X^{\prime})
\end{split}
\label{eq:ros}
\end{equation}

\begin{equation}
\begin{split}
\operatorname{RRS}\left(X, X^{\prime}, e_{X}, e_{X^{\prime}}\right) &= \max_{X^{\prime}} \frac{\left\|\frac{\left(e_{X}-e_{X^{\prime}}\right)}{e_{X}}\right\|_p}{\max \left(\left\|\frac{\left(\mathcal{L}_{X}-\mathcal{L}_{X^{\prime}}\right)}{\mathcal{L}_{X}}\right\|_p, \epsilon_{\min }\right)} \,,\\
&\quad \forall X^{\prime} \text{ s.t. } X^{\prime} \in \mathcal{N}_{X} ; f(X) = f(X^{\prime}) \,.
\end{split}
\label{eq:rrs}
\end{equation}

\subsection{CP Dataset}

The dataset used in this study originates from Groos et al.~\cite{groos2022development}, and is made up of 557 infants at elevated risk of perinatal brain injury, gathered from 13 different hospitals in Norway, Belgium, India, and the US. In compliance with Prechtl's GMA tool protocols, the infants were filmed during the FM period occurring between 9 and 18 weeks' corrected age. Using the CP decision tree from the Surveillance of Cerebral Palsy in Europe \cite{cans2000surveillance}, a pediatrician determined their CP statuses at 12 months corrected age or older. Of the 557 videos, 75\% were allocated for model training and validation, while the remaining 25\% formed the test set which is used in this paper.

Of the 139 videos in the test set, 21 are true CP cases and 118 are true No CP cases. For the XAI metrics testing, extracted skeleton tracker data from the original videos were used instead of the videos themselves. Given the long testing times for computing XAI metrics (approximately 1 hour per 5-second window on an RTX3090 GPU), random 5-second window samples were taken from the tracker data, proportional to the video length. For example, only one 5-second window sample was taken from the shortest videos.

Moreover, since testing for $stability$ metric requires that the true label matches the predicted label, only correctly predicted data were used, resulting in 15 CP and 111 No CP cases. From these, 24 random 5-second windows were selected from CP data and 136 from No CP data, maintaining the original CP to No-CP ratio. Additionally, windows at the beginning or end of the videos were avoided to reduce noise, and non-overlapping 5-second windows were chosen to avoid redundant data.

\subsection{CP Prediction Model}
\begin{figure*}[htbp]
    \centering
    \begin{subfigure}[t]{0.8\textwidth}
        \centering
        \includegraphics[width=\textwidth]{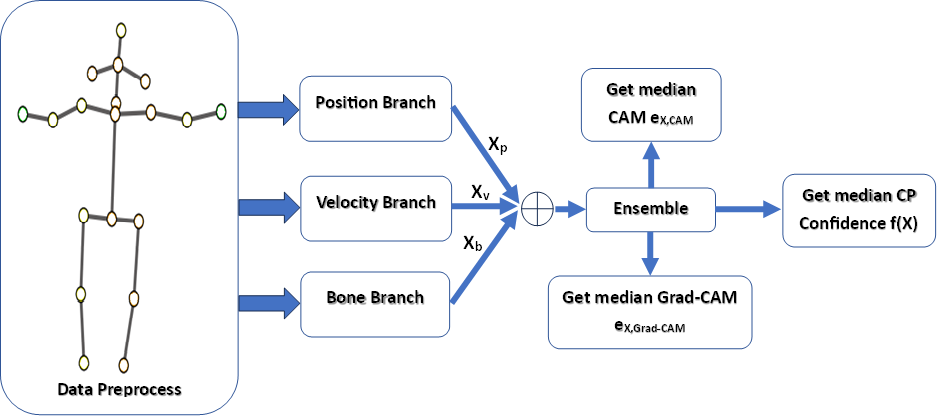}
        \caption{}
        \label{fig:flow}
    \end{subfigure}
    \vfill
    \begin{subfigure}[t]{0.8\textwidth}
        \centering
        \includegraphics[width=\textwidth]{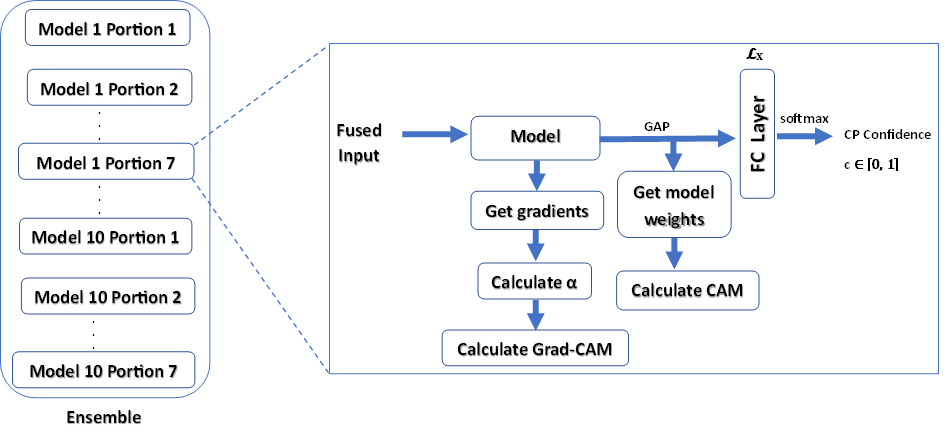}
        \caption{}
        \label{fig:ensemble}
    \end{subfigure}
    \caption{Overview of CP prediction ensemble pipeline, showing
    (\protect\subref{fig:flow})
    the variables for XAI metrics evaluation, and
    (\protect\subref{fig:ensemble})
    the processing steps for each model in the CP prediction ensemble.}
    \label{fig:methods}
\end{figure*}

The same deep learning-based CP prediction model from the Groos et al. study \cite{groos2022development} was evaluated for the XAI metrics. This model uses a GCN architecture to process infants' biomechanical movement properties, namely positions, velocities, and bones (distances between body keypoints). The GCN architecture was optimized through an automatic search, which created 70 distinct model instances, each trained on different sections of the dataset. Refinement of the model involved hyperparameter tuning and an automatic Neural Architecture Search (NAS) approach, exploring various architectural designs and configurations. The final Ensemble-NAS-GCN model combines predictions from the 70 GCN instances, as illustrated in Fig. \ref{fig:ensemble}, where the model number represents a unique architecture and the portion number denotes the specific training/validation fold used for that model. Details about each of the 10 GCN model architectures can be found in the Appendix Table \ref{tab:gcn_architectures}. Each model instance analyzes 5-second windows, with the CP risk determined as the median prediction across the ensemble. Similarly, for each 5-second window, the XAI attribution scores (i.e. CAM or Grad-CAM) from all 70 GCN instances were collected and unified by calculating the median score for each body point. There were no alterations made to the CP prediction pipeline except for capturing intermediate variables, such as model weights and gradients needed for the XAI metric evaluation. 
\begin{figure*}[htbp]
    \centering
    \includegraphics[width=0.8\textwidth]{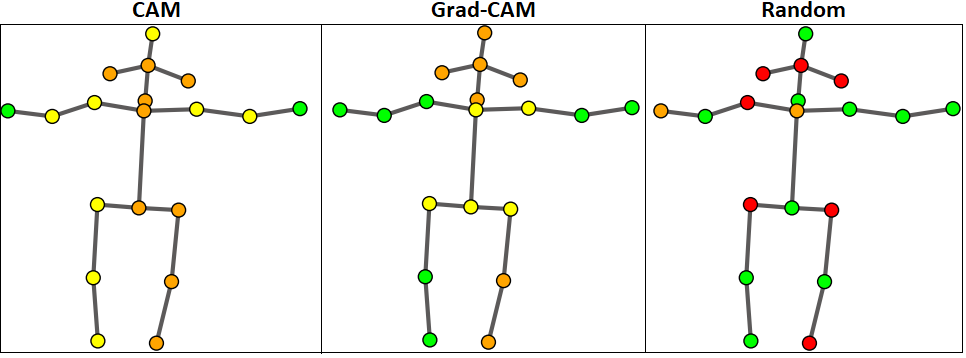}
    \caption{Sample visualization of attribution scores coming from the ensemble model's XAI methods tested using the same video. From left to right: CAM, Grad-CAM, and random method. Green indicates low attribution scores, yellow indicates moderate scores, orange shows high scores, and red signifies very high scores relative to the defined threshold score.}
    \label{fig:skeleton}
\end{figure*}
%%%%%%%%%%%%%%%%%%%%%%%%%%%%%%%%%%%%%%%%%%
\section{Methods}
\subsection{Skeleton Data Perturbation}

As demonstrated in our previous work \cite{pellano2024movements}, skeleton joints can be perturbed by converting Cartesian coordinates to spherical coordinates. This is performed using the equations below, where \(dx\) and \(dy\) represent the perturbation magnitudes along the $x$- and $y$-axes respectively, \(P\) is the original point, \(P'\) is the new perturbed point, \(x'\) and \(y'\) are the new coordinates, \(x\) and \(y\) are the original coordinates, and \(\theta\) is the randomly generated azimuthal angle:

\begin{align*}
r &= \|P - P'\| \\
dx &= r \cos\theta \\
x' &= \text{x} + dx \\
dy &= r \sin\theta \\
y' &= \text{y} + dy \,.
\end{align*}

Equations~\eqref{eq:ris},~\eqref{eq:ros},~\eqref{eq:rrs} require that the perturbed input data \(X'\) remains close to the original data \(X\), which also maintains accurate human kinematics and preserving the integrity of model predictions. To achieve this, \(r\) is set to 1\% of the median height of the infant across all frames in a 5-second window. The infant coordinate data are in pixels and height is calculated as the Euclidean distance between the head and left ankle. The values \(dx\) and \(dy\) are computed once for each joint and applied consistently across all video frames, producing a perturbed 2D point.

\subsection{Calculation and Evaluation of XAI Metrics}
\label{subsection:calculation}
The first step is to obtain the explanation for the original unperturbed data. The skeleton tracker data is fed into the CP prediction pipeline after \textit{Data Preprocess} stage as shown in Fig. \ref{fig:flow}. Each model in the ensemble generates its own explanation according to Equations \eqref{eq:cam} and \eqref{eq:gradcam}, as shown in Fig. \ref{fig:ensemble}). These individual explanations are combined by calculating the median values, representing the overall explanation for the ensemble. Similarly, individual model output predictions are combined by taking the median value, representing the ensemble's output prediction. This process yields the terms \( f(X) \), \( e_X \), and \( X \). The term \(\mathcal{L}_X\) is derived by collecting internal representations from the final FC layer before the softmax function of each model and flattening them into a single array.

Next, the body keypoints are ranked by importance from highest to lowest, and the top-k and non top-k joints are identified as determined by the ensemble's explanations. A sample visualization of these attribution scores, translated into color-coded importance indicators, is shown in Fig. \ref{fig:skeleton}, where green indicates low scores, yellow indicates moderate scores, orange indicates high scores, and red indicates very high scores relative to a threshold of 0.3. Perturbed varieties of the original data are generated as outlined in the previous section. Each 5-second window undergoes $n=50$ perturbations at a specified magnitude $r$. The top-k body points (where k = 1 to 19) are systematically perturbed $n$ times and then fed into the model to calculate PGI, then the remaining points are perturbed to compute PGU. Stability metrics are also computed using the explanations for the $n$ perturbations and the resulting intermediate values in the ensemble. We use the Area Under the Curve (AUC) to combine the calculated metric scores across all k values for each video into a single score. We then compute the mean and standard deviation of these AUCs in all videos to obtain the overall metric values. Since the metrics are unitless, the results are also compared against a random method that assigns feature attribution scores randomly, which represents the worst performance for an XAI method when subjected to a perturbation magnitude $r$.

To compare the ensemble's XAI metrics performance with individual models,  we conducted the same test on each model architecture. The ensemble consists of 70 model instances across 10 unique architectures, each trained on 7 different folds. Model 9, portion 5 achieved the highest AUC-ROC score in the test set, and thus portion 5 was used to represent each architecture in the tests. The same test pipeline is implemented as described above, except that the intermediate values for the metrics were derived from the individual models rather than the ensemble. Note that each model and the ensemble have unique sets of top-k and non top-k body points identified, resulting in distinct perturbed skeleton data, requiring separate tests. Using our hardware setup with RTX3090 GPU, the calculation for the ensemble took approximately 163 hours for CAM, 238 hours for Grad-CAM, and 120 hours for random method. The calculation for CAM, Grad-CAM, and random methods combined took approximately 24 hours for each model. The results for the individual models can be found in the Appendix Tables \ref{tab:metrics_results1} and \ref{tab:metrics_results2}. Lastly, we conducted an unpaired t-test to determine the statistical significance of the differences in metric results between CAM and Grad-CAM. For completeness, we also performed t-tests comparing CAM versus random attribution and Grad-CAM versus random attribution, expecting significant statistical differences in both tests.

\section{Results and Discussion}

\begin{figure*}[htbp]
    \centering
    \includegraphics[width=0.8\textwidth]{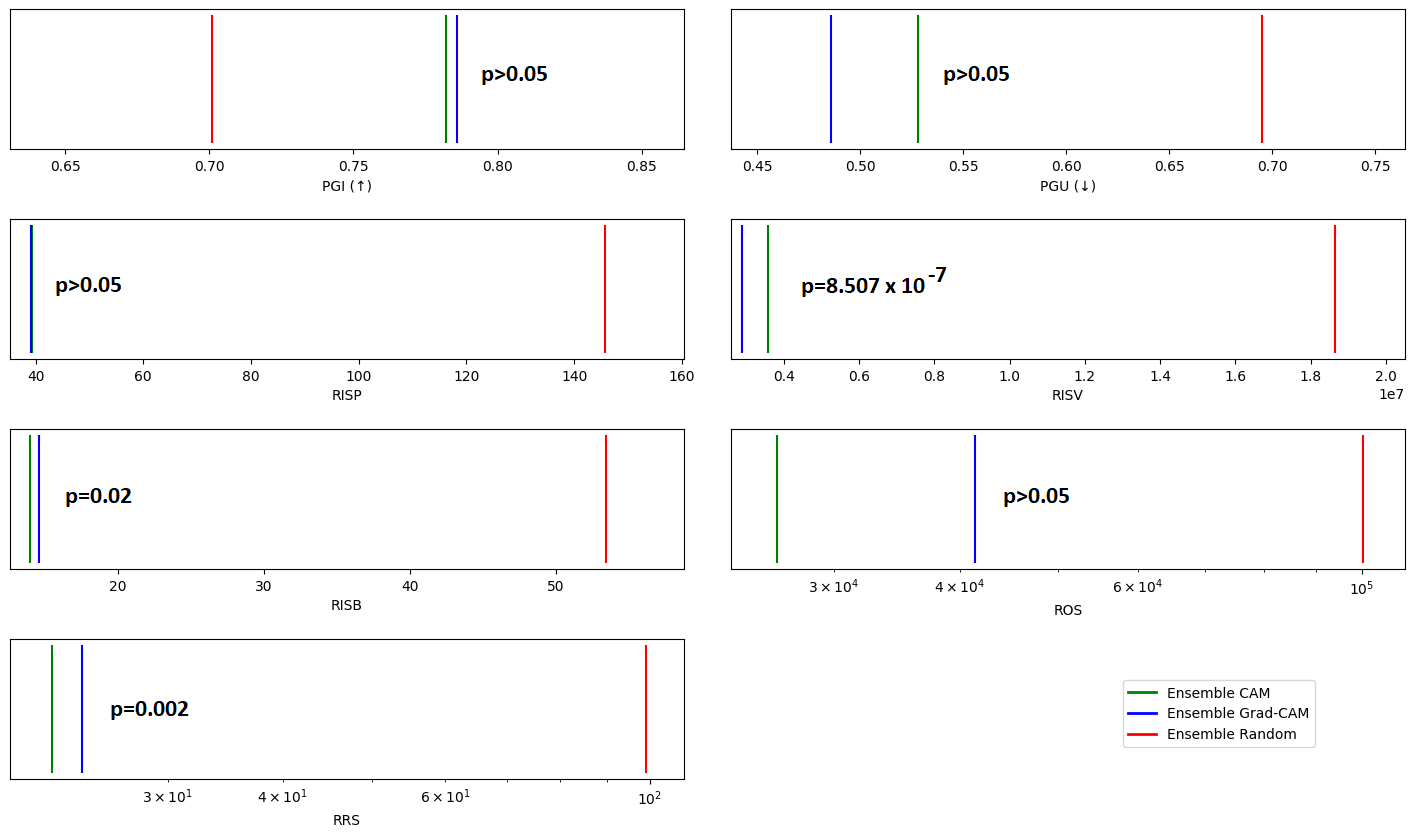}
    \caption{A line plot showing the metrics test results. PGI (↑) indicates that higher scores denote better performance, while PGU (↓) indicates that lower scores denote better performance. For stability metrics (RISP, RISV, RISB, ROS, RRS), values closer to zero are optimal. ROS and RRS are plotted on logarithmic scales to better display a wide range of values. Each plot also shows the p-value from the unpaired t-test between CAM and Grad-CAM for easier reference on the statistical significance of their difference.}
    \label{fig:results_ensemble}
\end{figure*}
A line plot is used to comparatively show the metrics performance of each XAI method in Fig. \ref{fig:results_ensemble}, which shows the relative performances for each metric based on their positions on the horizontal axis. The exact numerical results with confidence intervals are shown in the Appendix Tables \ref{tab:metrics_results1} and \ref{tab:metrics_results2}.

\subsection{Faithfulness}
PGI results indicate that the ensemble’s predictions are influenced by key features identified by the two XAI methods (i.e. the change in output probability $f(X)$ increases when important nodes are perturbed), with a significant difference compared to the random attribution at $p < 0.005$ for both CAM versus random and Grad-CAM versus random. There was no significant difference in PGI and PGU between CAM and Grad-CAM. Overall, the faithfulness tests indicate that XAI methods effectively distinguish important from unimportant body points influencing predictions.

\subsection{Stability}
The plots of RISp, RISv, RISb, ROS, and RRS in Fig. \ref{fig:results_ensemble} immediately show us that random explanations cause significant changes in the stability results, with $p < 0.005$ for both CAM versus random and Grad-CAM versus random. This indicates that both XAI methods are less susceptible to large explanation changes with small input perturbations. Most notably, Grad-CAM offers a highly significant improvement (with $p < 8.507 \times 10^{-7}$) of input stability compared to CAM for RISv of the ensamble model. In contrast, CAM offers a significantly better input stability (with $p < 0.05$) compared to Grad-CAM for RISb and RRS. The result for the stability and faithfullness metric for the different GCN models in ensemble can be found in the Appendix Tables \ref{tab:metrics_results1} and \ref{tab:metrics_results2}.

%%%%%%%%%%%%%%%%%%%%%%%%%%%%%%%%%%%%%%%%%%
\section{Conclusion and Recommendation}
This study evaluated the reliability and robustness of two XAI techniques, CAM and Grad-CAM, within a deep learning ensemble model. The faithfulness metrics show both methods are effective in identifying important and unimportant body points influencing CP predictions, while stability tests demonstrate robustness against minor data perturbations. Specifically, CAM significantly outperformed Grad-CAM in RISb and RRS, while Grad-CAM excelled in RISv. The choice of XAI method depends on the specific application and key metrics. For instance, to identify potential movement biomarkers for CP related to joint velocity, Grad-CAM is recommended for its superior RISv performance. This also suggests exploring the velocity input branch for possible CP movement biomarkers, which aligns with previous studies \cite{meinecke2006movement, adde2009using, adde2010early, stahl2012optical} where a velocity parameter is used in classical machine learning assessments of CP. If similar explanations for body points are found in multiple 5-second windows, Grad-CAM offers better insight into body point velocity due to its stable explanations despite input variations. Conversely, for quick explanations (i.e. for visual inspection purposes), CAM is preferable due to its faster calculation time.

Individual models in the ensemble show varied results in metrics tests, while the ensemble provides a combined outcome from its constituent models. In our previous work \cite{pellano2024movements}, we applied metrics tests to a single GCN-based model, limiting generalizability. This study has evaluated XAI metrics in both an ensemble and the individual models composing it, revealing that neither CAM nor Grad-CAM consistently outperforms the other in all metrics. Overall, our findings demonstrate that both XAI methods can provide reliable and stable explanations in CP prediction models. By finding patterns and doing further studies into the explanations, these could potentially supplement GMA observers with specialized domain knowledge, contributing to more interpretable and trustworthy AI diagnostics. The study's insights into the comparative performance of CAM and Grad-CAM can also guide future research in improving AI explainability in other high-stakes medical applications. As a next step, it is crucial to introduce more XAI metrics that incorporate domain-specific knowledge to further validate the relevance and accuracy of the explanations provided. It is also necessary to perform clinical interpretations on the explanations to enhance the practical utility of the models.

%\section{REFERENCES}
\bibliographystyle{IEEEtran}
%\bibliography{ref}

% References should be produced using the bibtex program from suitable
% BiBTeX files (here: strings, refs, manuals). The IEEEbib.bst bibliography
% style file from IEEE produces unsorted bibliography list.
% -------------------------------------------------------------------------
%\bibliographystyle{IEEEbib}
%\bibliography{strings,refs}
%\onecolumn
\appendix
%\section{}
%\label{appendix:class11data}

%\section{}
%\label{appendix:class26data}

\begin{table*}[h!]
\centering
\caption{Metrics Test Results for Ensemble and Individual Models (part 1)}
\small
\begin{tabular}{|l|c|c|c|}
\hline
\textbf{PGI} & \textbf{CAM} & \textbf{Grad-CAM} & \textbf{Random} \\
\hline
\textbf{Ensemble} & 0.782 ± 0.141 & 0.786 ± 0.137 & 0.701 ± 0.130 \\
\textbf{Model 1} & 1.086 ± 0.175 & 1.088 ± 0.174 & 0.971 ± 0.149 \\
\textbf{Model 2} & 1.379 ± 0.264 & 1.359 ± 0.263 & 1.132 ± 0.234 \\
\textbf{Model 3} & 1.305 ± 0.295 & 1.303 ± 0.296 & 1.179 ± 0.280 \\
\textbf{Model 4} & 1.085 ± 0.244 & 0.976 ± 0.211 & 0.909 ± 0.199 \\
\textbf{Model 5} & 1.516 ± 0.315 & 1.535 ± 0.315 & 1.234 ± 0.284 \\
\textbf{Model 6} & 1.231 ± 0.218 & 1.247 ± 0.222 & 1.146 ± 0.200 \\
\textbf{Model 7} & 0.711 ± 0.216 & 0.739 ± 0.209 & 0.623 ± 0.168 \\
\textbf{Model 8} & 1.523 ± 0.181 & 1.521 ± 0.181 & 1.378 ± 0.157 \\
\textbf{Model 9} & 1.048 ± 0.193 & 1.333 ± 0.253 & 1.053 ± 0.196 \\
\textbf{Model 10} & 0.680 ± 0.224 & 0.621 ± 0.208 & 0.762 ± 0.283 \\
\hline
\textbf{PGU} & \textbf{CAM} & \textbf{Grad-CAM} & \textbf{Random} \\
\hline
\textbf{Ensemble} & 0.528 ± 0.095 & 0.486 ± 0.103 & 0.695 ± 0.128 \\
\textbf{Model 1} & 0.818 ± 0.152 & 0.817 ± 0.152 & 1.063 ± 0.172 \\
\textbf{Model 2} & 0.686 ± 0.138 & 0.683 ± 0.138 & 0.918 ± 0.182 \\
\textbf{Model 3} & 0.946 ± 0.242 & 0.939 ± 0.242 & 1.162 ± 0.279 \\
\textbf{Model 4} & 0.543 ± 0.105 & 0.633 ± 0.130 & 0.885 ± 0.184 \\
\textbf{Model 5} & 0.676 ± 0.150 & 0.680 ± 0.150 & 1.066 ± 0.197 \\
\textbf{Model 6} & 0.990 ± 0.179 & 0.990 ± 0.179 & 1.032 ± 0.200 \\
\textbf{Model 7} & 0.613 ± 0.175 & 0.534 ± 0.146 & 0.624 ± 0.172 \\
\textbf{Model 8} & 1.170 ± 0.154 & 1.169 ± 0.154 & 1.339 ± 0.165 \\
\textbf{Model 9} & 1.028 ± 0.218 & 0.754 ± 0.148 & 0.981 ± 0.188 \\
\textbf{Model 10} & 0.525 ± 0.173 & 0.628 ± 0.209 & 0.588 ± 0.178 \\
\hline
\textbf{RISp} & \textbf{CAM} & \textbf{Grad-CAM} & \textbf{Random} \\
\hline
\textbf{Ensemble} & 39.328 ± 3.257 & 39.167 ± 2.544 & 145.805 ± 4.371 \\
\textbf{Model 1} & 48.779 ± 3.083 & 48.779 ± 3.083 & 162.523 ± 4.652 \\
\textbf{Model 2} & 37.047 ± 2.456 & 37.055 ± 2.458 & 162.542 ± 5.308 \\
\textbf{Model 3} & 40.152 ± 2.474 & 39.874 ± 2.457 & 161.785 ± 4.371 \\
\textbf{Model 4} & 33.725 ± 1.730 & 34.488 ± 3.546 & 161.706 ± 4.784 \\
\textbf{Model 5} & 30.154 ± 2.790 & 30.154 ± 2.790 & 162.705 ± 4.718 \\
\textbf{Model 6} & 37.997 ± 4.055 & 38.262 ± 4.056 & 162.625 ± 5.045 \\
\textbf{Model 7} & 36.014 ± 2.067 & 37.290 ± 3.642 & 165.739 ± 4.367 \\
\textbf{Model 8} & 49.747 ± 2.182 & 49.751 ± 2.181 & 157.211 ± 5.326 \\
\textbf{Model 9} & 37.048 ± 2.160 & 26.384 ± 3.928 & 162.735 ± 4.865 \\
\textbf{Model 10} & 35.644 ± 3.336 & 53.884 ± 3.070 & 160.971 ± 5.069 \\
\hline
\textbf{RISv} & \textbf{CAM} & \textbf{Grad-CAM} & \textbf{Random} \\
\hline
\textbf{Ensemble} & 3585402.774 ± 544958.663 & 2888248.714 ± 486164.011 & 18641327.196 ± 3924550.555 \\
\textbf{Model 1} & 5013896.473 ± 733566.854 & 5013897.457 ± 733567.024 & 22808539.350 ± 6802897.522 \\
\textbf{Model 2} & 3826587.193 ± 1326590.444 & 3842761.467 ± 1328900.588 & 22426153.524 ± 4301241.358 \\
\textbf{Model 3} & 3025573.782 ± 581141.137 & 3021450.227 ± 581182.591 & 21338036.160 ± 5862771.374 \\
\textbf{Model 4} & 3756392.648 ± 438413.881 & 3566904.056 ± 719258.913 & 21367999.081 ± 4543133.766 \\
\textbf{Model 5} & 2271316.786 ± 577805.929 & 2271316.654 ± 577805.925 & 22637633.435 ± 4817636.850 \\
\textbf{Model 6} & 5008019.480 ± 868948.611 & 5098120.204 ± 882587.754 & 22424391.642 ± 5949317.869 \\
\textbf{Model 7} & 4669245.654 ± 571611.191 & 4169000.548 ± 1686004.537 & 23354361.639 ± 5345484.607 \\
\textbf{Model 8} & 5592108.186 ± 760098.099 & 5586872.125 ± 760912.793 & 19221557.170 ± 3435013.322 \\
\textbf{Model 9} & 5011561.220 ± 1588484.050 & 2193192.993 ± 951410.356 & 22146716.652 ± 6454643.229 \\
\textbf{Model 10} & 3480586.455 ± 669467.413 & 6846223.738 ± 2223237.831 & 22937902.126 ± 5540543.868 \\
\hline
\end{tabular}
\label{tab:metrics_results1}
\end{table*}

\begin{table*}[h!]
\centering
\caption{Metrics Test Results for Ensemble and Individual Models (part 2)}
\small
\begin{tabular}{|l|c|c|c|}
\hline
\textbf{RISb} & \textbf{CAM} & \textbf{Grad-CAM} & \textbf{Random} \\
\hline
\textbf{Ensemble} & 13.977 ± 1.240 & 14.575 ± 0.992 & 53.456 ± 1.641 \\
\textbf{Model 1} & 18.565 ± 1.301 & 18.565 ± 1.301 & 59.152 ± 1.826 \\
\textbf{Model 2} & 12.941 ± 0.917 & 12.970 ± 0.924 & 60.379 ± 2.114 \\
\textbf{Model 3} & 14.947 ± 0.944 & 14.909 ± 0.943 & 59.725 ± 1.783 \\
\textbf{Model 4} & 12.904 ± 0.687 & 13.256 ± 1.360 & 61.108 ± 1.750 \\
\textbf{Model 5} & 11.153 ± 1.096 & 11.154 ± 1.096 & 60.242 ± 1.856 \\
\textbf{Model 6} & 12.638 ± 1.369 & 12.698 ± 1.366 & 60.179 ± 2.090 \\
\textbf{Model 7} & 12.220 ± 0.823 & 12.921 ± 1.298 & 61.857 ± 1.852 \\
\textbf{Model 8} & 18.619 ± 0.839 & 18.606 ± 0.841 & 58.860 ± 2.132 \\
\textbf{Model 9} & 13.195 ± 0.815 & 9.910 ± 1.501 & 59.681 ± 1.949 \\
\textbf{Model 10} & 13.266 ± 1.256 & 19.019 ± 1.077 & 59.780 ± 1.875 \\
\hline
\textbf{ROS} & \textbf{CAM} & \textbf{Grad-CAM} & \textbf{Random} \\
\hline
\textbf{Ensemble} & 26378.525 ± 22364.341 & 41400.726 ± 40902.013 & 100282.533 ± 45148.804 \\
\textbf{Model 1} & 12358.423 ± 5107.679 & 12358.425 ± 5107.678 & 167488.094 ± 188907.518 \\
\textbf{Model 2} & 48508.845 ± 46128.105 & 48507.195 ± 46128.132 & 238020.249 ± 149941.173 \\
\textbf{Model 3} & 689995.736 ± 563561.376 & 688867.883 ± 563590.546 & 17526781.455 ± 24820904.705 \\
\textbf{Model 4} & 625495.413 ± 539095.851 & 616604.749 ± 730402.984 & 5222686.783 ± 4820274.414 \\
\textbf{Model 5} & 4888.137 ± 2348.803 & 4888.130 ± 2348.801 & 188705.842 ± 96210.421 \\
\textbf{Model 6} & 18761.681 ± 16394.279 & 11717.707 ± 7851.360 & 88558.495 ± 38702.627 \\
\textbf{Model 7} & 30960887.120 ± 49570977.534 & 2465358.603 ± 1545351.239 & 68234634.236 ± 65663999.020 \\
\textbf{Model 8} & 11131.461 ± 9793.839 & 11131.035 ± 9793.876 & 31365.712 ± 39137.823 \\
\textbf{Model 9} & 75462.505 ± 51427.097 & 30556.754 ± 23305.412 & 272482.338 ± 174124.522 \\
\textbf{Model 10} & 18182708708.536 ± 34033740331.051 & 3903058531927.605 ± 7702513374782.742 & 78180252904.944 ± 138655149933.328 \\
\hline
\textbf{RRS} & \textbf{CAM} & \textbf{Grad-CAM} & \textbf{Random} \\
\hline
\textbf{Ensemble} & 22.482 ± 2.343 & 24.196 ± 1.870 & 99.062 ± 6.989 \\
\textbf{Model 1} & 0.577 ± 0.290 & 0.577 ± 0.290 & 1.661 ± 0.726 \\
\textbf{Model 2} & 0.189 ± 0.138 & 0.189 ± 0.138 & 0.619 ± 0.478 \\
\textbf{Model 3} & 0.484 ± 0.300 & 0.484 ± 0.300 & 1.382 ± 0.708 \\
\textbf{Model 4} & 0.168 ± 0.117 & 0.198 ± 0.205 & 0.985 ± 0.547 \\
\textbf{Model 5} & 0.443 ± 0.252 & 0.443 ± 0.252 & 1.356 ± 0.729 \\
\textbf{Model 6} & 0.364 ± 0.242 & 0.417 ± 0.270 & 1.460 ± 0.753 \\
\textbf{Model 7} & 0.198 ± 0.165 & 0.236 ± 0.198 & 0.627 ± 0.479 \\
\textbf{Model 8} & 0.984 ± 0.401 & 0.984 ± 0.401 & 2.196 ± 0.879 \\
\textbf{Model 9} & 0.506 ± 0.241 & 0.268 ± 0.183 & 1.570 ± 0.747 \\
\textbf{Model 10} & 0.229 ± 0.199 & 0.347 ± 0.263 & 0.819 ± 0.581 \\
\hline
\end{tabular}
\label{tab:metrics_results2}
\end{table*}

\begin{table*}[h!]
\centering
\caption{The architectures of the 10 models in the ensemble. Abbreviations: br., branch; bottl., bottleneck; MBC., mobile inverted bottleneck convolution; den., dense; mod., module; tmp., temporal; lin., linear; gl., global; sp., spatial; DA, disentangled aggregation; SC, spatial configuration; SE, Squeeze-and-Excitation; abs., absolute; ch., channel; ReLU, rectified linear unit; sw., swish; AUC, area under the receiver operating characteristic curve.}
\small
\begin{adjustbox}{max width=\textwidth}
\begin{tabular}{|>{\centering\arraybackslash}m{5.5cm}|>{\centering\arraybackslash}m{0.6cm}|>{\centering\arraybackslash}m{0.6cm}|>{\centering\arraybackslash}m{0.6cm}|>{\centering\arraybackslash}m{0.6cm}|>{\centering\arraybackslash}m{0.6cm}|>{\centering\arraybackslash}m{0.6cm}|>{\centering\arraybackslash}m{0.6cm}|>{\centering\arraybackslash}m{0.6cm}|>{\centering\arraybackslash}m{0.6cm}|>{\centering\arraybackslash}m{0.6cm}|}
\hline
\textbf{Architectural choice} & 1 & 2 & 3 & 4 & 5 & 6 & 7 & 8 & 9 & 10 \\
\hline
No. modules of input br. & 3 & 3 & 2 & 2 & 2 & 2 & 3 & 3 & 2 & 2 \\
Width of input br. & 10 & 10 & 12 & 10 & 8 & 6 & 8 & 6 & 12 & 12 \\
Block type in initial mod. & Bottl. & Basic & Basic & Basic & Bottl. & Basic & Basic & MBC. & Bottl. & Basic \\
Residual type in initial mod. & None & Den. & None & Block & Den. & Den. & Mod. & Block & Den. & Den. \\
No. tmp. scales in input br. & 1 & 3 & 2 & 2 & 3 & 2 & 2 & 1 & 2 & 2 \\
No. levels of main br. & 3 & 1 & 3 & 2 & 2 & 2 & 2 & 2 & 2 & 1 \\
No. modules of main br. levels & 3 & 2 & 1 & 1 & 3 & 3 & 2 & 3 & 1 & 1 \\
Width of first level of main br. & 12 & 12 & 12 & 10 & 12 & 12 & 10 & 12 & 8 & 12 \\
No. tmp. scales in main br. & 1 & 2 & 2 & 3 & 2 & 1 & 2 & 1 & 1 & 1 \\
Pooling layer type & Gl. & Gl. & Gl. & Sp. & Sp. & Gl. & Gl. & Gl. & Gl. & Sp. \\
Graph convolution type & DA 2 & DA 4+2 & SC & DA 4 & SC & DA 4 & DA 2 & DA 2 & DA 4 & SC \\
Block type & Basic & MBC. & Basic & Basic & Bottl. & Bottl. & Basic & Basic & Bottl. & Basic \\
Bottl. factor & 4 & 2 & 2 & 2 & 2 & 2 & 4 & 4 & 4 & 4 \\
Residual type & None & Block & Den. & None & None & Block & None & Den. & Block & None \\
SE type & None & Outer & Inner & None & Outer & None & None & Outer & Outer & None \\
SE ratio & - & 2 & 2 & 4 & 2 & 4 & 4 & 4 & 4 & 4 \\
SE ratio type & - & Abs. & Abs. & Abs. & - & Abs. & - & Abs. & Abs. & - \\
Attention type & Ch. & - & - & - & Ch. & - & Ch. & - & Ch. & - \\
Nonlinearity type & ReLU & Sw. &  ReLU & Sw. & Sw. & ReLU & ReLU & Sw. & ReLU & Sw. \\
Tmp. kernel size & 9 & 7 & 9 & 7 & 9 & 7 & 9 & 7 & 9 & 7 \\
AUC & 0.949 & 0.942 & 0.938 & 0.943 & 0.937 & 0.956 & 0.953 & 0.953 & 0.932 & 0.947 \\
\hline
\end{tabular}
\label{tab:gcn_architectures}
\end{adjustbox}
\end{table*}

\begin{table}[ht]
\centering
\caption{Unpaired t-test result: CAM versus Grad-CAM (part 1)}
\small
\begin{tabular}{lcc}
\toprule
\multicolumn{3}{c}{PGI} \\
\midrule
 & t-statistic & p-value \\
\midrule
Ensemble & 0.346594 & 0.728908  \\
Model 1 & -0.059377 & 0.952654  \\
Model 2 & 0.402978 & 0.686979  \\
Model 3 & 0.038596 & 0.969214  \\
Model 4 & 2.602390 & 0.009280  \\
Model 5 & -0.316427 & 0.751690  \\
Model 6 & -0.382246 & 0.702292  \\
Model 7 & -0.592449 & 0.553572  \\
Model 8 & 0.063267 & 0.949556  \\
Model 9 & -5.950216 & 2.832418e-09  \\
Model 10 & 1.353371 & 0.175988 \\
\midrule
\multicolumn{3}{c}{PGU} \\
\midrule
 & t-statistic & p-value \\
\midrule
Ensemble & -1.908225 & 0.056412 \\
Model 1 & 0.015310 & 0.987785 \\
Model 2 & 0.092156 & 0.926577 \\
Model 3 & 0.141020 & 0.887859 \\
Model 4 & -3.460206 & 0.000544 \\
Model 5 & -0.136616 & 0.891339 \\
Model 6 & -0.018139 & 0.985529 \\
Model 7 & 2.238301 & 0.025240 \\
Model 8 & 0.019318 & 0.984588 \\
Model 9 & 6.835406 & 9.093938e-12 \\
Model 10 & -2.544099 & 0.010982 \\
\midrule
\multicolumn{3}{c}{RISp} \\
\midrule
 & t-statistic & p-value \\
\midrule
Ensemble & -0.437441 & 0.661808 \\
Model 1 & -0.000016 & 0.999988 \\
Model 2 & -0.024395 & 0.980538 \\
Model 3 & 0.493987 & 0.621334 \\
Model 4 & -1.033182 & 0.301573 \\
Model 5 & -0.000004 & 0.999997 \\
Model 6 & -0.183744 & 0.854220 \\
Model 7 & -2.225563 & 0.026086  \\
Model 8 & -0.002316 & 0.998152 \\
Model 9 & 15.424594 & 1.675449e-52 \\
Model 10 & -24.577841 & 3.367287e-127 \\
\midrule
\multicolumn{3}{c}{RISv} \\
\midrule
 & t-statistic & p-value \\
\midrule
Ensemble & -4.928563 & 8.507249e-07 \\
Model 1 & -0.000004 & 0.999996 \\
Model 2 & -0.057351 & 0.954267 \\
Model 3 & 0.025330 & 0.979793 \\
Model 4 & 1.038284 & 0.299183 \\
Model 5 & 0.000001 & 0.999999 \\
Model 6 & -0.372872 & 0.709257 \\
Model 7 & 1.580504 & 0.114070 \\
Model 8 & 0.023271 & 0.981435 \\
Model 9 & 8.550714 & 1.555406e-17 \\
Model 10 & -8.233855 & 2.512258e-16 \\
\bottomrule
\end{tabular}
\label{tab:t-test1}
\end{table}

\begin{table}[ht]
\centering
\caption{Unpaired t-test result: CAM versus Grad-CAM (part 2)}
\small
\begin{tabular}{lcc}
\toprule
\multicolumn{3}{c}{RISb} \\
\midrule
 & t-statistic \\
\midrule
Ensemble & 2.313236 & 0.020745 \\
Model 1 & -0.000016 & 0.999987 \\
Model 2 & -0.145161 & 0.884588 \\
Model 3 & 0.167647 & 0.866867 \\
Model 4 & -1.296128 & 0.194995 \\
Model 5 & -0.000001 & 0.999999 \\
Model 6 & -0.105578 & 0.915920 \\
Model 7 & -3.081867 & 0.002068 \\
Model 8 & 0.067254 & 0.946382 \\
Model 9 & 12.590592 & 8.358464e-36 \\
Model 10 & -20.963022 & 3.321069e-94 \\
\midrule
\multicolumn{3}{c}{ROS} \\
\midrule
 & t-statistic & p-value \\
\midrule
Ensemble & 0.852538 & 0.393952 \\
Model 1 & -0.000001 & 1.000000 \\
Model 2 & -0.000000 & 1.000000 \\
Model 3 & 0.004623 & 0.996312 \\
Model 4 & 0.053704 & 0.957173 \\
Model 5 & 0.000005 & 0.999996 \\
Model 6 & 0.883619 & 0.376956 \\
Model 7 & 3.327644 & 0.000886 \\
Model 8 & 0.000058 & 0.999954 \\
Model 9 & 2.395027 & 0.016667 \\
Model 10 & -1.002149 & 0.316352 \\
\midrule
\multicolumn{3}{c}{RRS} \\
\midrule
 & t-statistic & p-value \\
\midrule
Ensemble & 3.068542 & 0.002161 \\
Model 1 & -0.000000 & 1.000000 \\
Model 2 & -0.000000 & 1.000000 \\
Model 3 & -0.000000 & 1.000000 \\
Model 4 & -0.000000 & 1.000000 \\
Model 5 & 0.000000 & 1.000000 \\
Model 6 & -1.152599 & 0.249122 \\
Model 7 & -0.914333 & 0.360580 \\
Model 8 & -0.000002 & 0.999998 \\
Model 9 & 5.372476 & 8.077018e-08 \\
Model 10 & -2.785957 & 0.005355 \\
\bottomrule
\end{tabular}
\label{tab:t-test2}
\end{table}

\begin{table}[ht]
\centering
\caption{Unpaired t-test result: CAM versus Random Attribution (part 1)}
\small
\begin{tabular}{lcc}
\toprule
\multicolumn{3}{c}{PGI} \\
\midrule
 & t-statistic & p-value \\
\midrule
Ensemble & 3.228758 & 0.001250 \\
Model 1 & 3.305932 & 0.000952 \\
Model 2 & 4.870700 & 0.000001 \\
Model 3 & 2.086048 & 0.037016 \\
Model 4 & 4.119581 & 0.000038 \\
Model 5 & 4.927732 & 0.000001 \\
Model 6 & 2.081000 & 0.037476 \\
Model 7 & 2.117884 & 0.034226 \\
Model 8 & 4.061314 & 0.000049 \\
Model 9 & -0.138202 & 0.890085 \\
Model 10 & -1.668719 & 0.095226 \\
\midrule
\multicolumn{3}{c}{PGU} \\
\midrule
 & t-statistic & p-value \\
\midrule
Ensemble & -8.240703 & 0.000000 \\
Model 1 & -6.558054 & 0.000000 \\
Model 2 & -6.145350 & 0.000000 \\
Model 3 & -4.001890 & 0.000064 \\
Model 4 & -10.571248 & 0.000000 \\
Model 5 & -9.798438 & 0.000000 \\
Model 6 & -1.126884 & 0.259839 \\
Model 7 & -0.319122 & 0.749646 \\
Model 8 & -4.784537 & 0.000002 \\
Model 9 & 1.025866 & 0.304999 \\
Model 10 & -1.681898 & 0.092643 \\
\midrule
\multicolumn{3}{c}{RISp} \\
\midrule
 & t-statistic & p-value \\
\midrule
Ensemble & -69.274966 & 0.000000 \\
Model 1 & -65.288282 & 0.000000 \\
Model 2 & -72.948463 & 0.000000 \\
Model 3 & -71.029458 & 0.000000 \\
Model 4 & -75.880358 & 0.000000 \\
Model 5 & -76.810741 & 0.000000 \\
Model 6 & -68.599757 & 0.000000 \\
Model 7 & -75.154906 & 0.000000 \\
Model 8 & -63.345906 & 0.000000 \\
Model 9 & -72.521749 & 0.000000 \\
Model 10 & -71.929170 & 0.000000 \\
\midrule
\multicolumn{3}{c}{RISv} \\
\midrule
 & t-statistic & p-value \\
\midrule
Ensemble & -24.344140 & 0.000000 \\
Model 1 & -20.114345 & 0.000000 \\
Model 2 & -24.965946 & 0.000000 \\
Model 3 & -21.361852 & 0.000000 \\
Model 4 & -23.671184 & 0.000000 \\
Model 5 & -26.727520 & 0.000000 \\
Model 6 & -20.797096 & 0.000000 \\
Model 7 & -24.131390 & 0.000000 \\
Model 8 & -21.404787 & 0.000000 \\
Model 9 & -19.301737 & 0.000000 \\
Model 10 & -24.397185 & 0.000000 \\
\bottomrule
\end{tabular}
\label{tab:t-test-camvsrandom1}
\end{table}

\begin{table}[ht]
\centering
\caption{Unpaired t-test result: CAM versus Random Attribution (part 2)}
\small
\begin{tabular}{lcc}
\toprule
\multicolumn{3}{c}{RISb} \\
\midrule
 & t-statistic & p-value \\
\midrule
Ensemble & -64.755147 & 0.000000 \\
Model 1 & -61.762571 & 0.000000 \\
Model 2 & -72.323794 & 0.000000 \\
Model 3 & -68.914145 & 0.000000 \\
Model 4 & -74.410386 & 0.000000 \\
Model 5 & -75.631103 & 0.000000 \\
Model 6 & -69.824165 & 0.000000 \\
Model 7 & -75.605835 & 0.000000 \\
Model 8 & -62.847825 & 0.000000 \\
Model 9 & -71.187557 & 0.000000 \\
Model 10 & -72.411265 & 0.000000 \\
\midrule
\multicolumn{3}{c}{ROS} \\
\midrule
 & t-statistic & p-value \\
\midrule
Ensemble & -3.342298 & 0.000836 \\
Model 1 & -1.645748 & 0.099920 \\
Model 2 & -3.598124 & 0.000325 \\
Model 3 & -2.312361 & 0.020826 \\
Model 4 & -2.644420 & 0.008225 \\
Model 5 & -4.820705 & 0.000001 \\
Model 6 & -4.349855 & 0.000014 \\
Model 7 & -1.872147 & 0.061255 \\
Model 8 & -1.033692 & 0.301357 \\
Model 9 & -3.696956 & 0.000221 \\
Model 10 & -1.496006 & 0.134750 \\
\midrule
\multicolumn{3}{c}{RRS} \\
\midrule
 & t-statistic & p-value \\
\midrule
Ensemble & -39.979699 & 0.000000 \\
Model 1 & -9.543686 & 0.000000 \\
Model 2 & -6.262358 & 0.000000 \\
Model 3 & -8.260879 & 0.000000 \\
Model 4 & -9.619116 & 0.000000 \\
Model 5 & -8.938158 & 0.000000 \\
Model 6 & -10.515924 & 0.000000 \\
Model 7 & -6.153502 & 0.000000 \\
Model 8 & -9.235283 & 0.000000 \\
Model 9 & -9.842633 & 0.000000 \\
Model 10 & -7.617952 & 0.000000 \\
\bottomrule
\end{tabular}
\label{tab:t-test-camvsrandom2}
\end{table}

\begin{table}[ht]
\centering
\caption{Unpaired t-test result: Grad-CAM versus Random Attribution (part 1)}
\small
\begin{tabular}{lcc}
\toprule
\multicolumn{3}{c}{PGI} \\
\midrule
 & t-statistic & p-value \\
\midrule
Ensemble & 2.839915 & 0.004528 \\
Model 1 & 3.369167 & 0.000759 \\
Model 2 & 4.476108 & 0.000008 \\
Model 3 & 2.046355 & 0.040764 \\
Model 4 & 1.475531 & 0.140122 \\
Model 5 & 5.247443 & 0.000000 \\
Model 6 & 2.441880 & 0.014639 \\
Model 7 & 2.781726 & 0.005424 \\
Model 8 & 3.995221 & 0.000065 \\
Model 9 & 5.820300 & 0.000000 \\
Model 10 & -2.938911 & 0.003307 \\
\midrule
\multicolumn{3}{c}{PGU} \\
\midrule
 & t-statistic & p-value \\
\midrule
Ensemble & -6.411041 & 0.000000 \\
Model 1 & -6.572715 & 0.000000 \\
Model 2 & -6.228821 & 0.000000 \\
Model 3 & -4.133470 & 0.000036 \\
Model 4 & -7.212538 & 0.000000 \\
Model 5 & -9.669277 & 0.000000 \\
Model 6 & -1.105639 & 0.268929 \\
Model 7 & -2.531826 & 0.011374 \\
Model 8 & -4.802966 & 0.000002 \\
Model 9 & -6.146440 & 0.000000 \\
Model 10 & 0.920952 & 0.357114 \\
\midrule
\multicolumn{3}{c}{RISp} \\
\midrule
 & t-statistic & p-value \\
\midrule
Ensemble & -67.438628 & 0.000000 \\
Model 1 & -65.288262 & 0.000000 \\
Model 2 & -72.931395 & 0.000000 \\
Model 3 & -71.205931 & 0.000000 \\
Model 4 & -73.143467 & 0.000000 \\
Model 5 & -76.810740 & 0.000000 \\
Model 6 & -68.494628 & 0.000000 \\
Model 7 & -72.465331 & 0.000000 \\
Model 8 & -63.339157 & 0.000000 \\
Model 9 & -76.109905 & 0.000000 \\
Model 10 & -61.907950 & 0.000000 \\
\midrule
\multicolumn{3}{c}{RISv} \\
\midrule
 & t-statistic & p-value \\
\midrule
Ensemble & -23.174673 & 0.000000 \\
Model 1 & -20.114344 & 0.000000 \\
Model 2 & -24.940226 & 0.000000 \\
Model 3 & -21.366473 & 0.000000 \\
Model 4 & -23.743872 & 0.000000 \\
Model 5 & -26.727520 & 0.000000 \\
Model 6 & -20.680862 & 0.000000 \\
Model 7 & -23.590898 & 0.000000 \\
Model 8 & -21.412311 & 0.000000 \\
Model 9 & -22.667079 & 0.000000 \\
Model 10 & -18.506366 & 0.000000 \\
\bottomrule
\end{tabular}
\label{tab:t-test-gcamvsrandom1}
\end{table}

\begin{table}[ht]
\centering
\caption{Unpaired t-test result: Grad-CAM versus Random Attribution (part 2)}
\small
\begin{tabular}{lcc}
\toprule
\multicolumn{3}{c}{RISb} \\
\midrule
 & t-statistic & p-value \\
\midrule
Ensemble & -64.493624 & 0.000000 \\
Model 1 & -61.762544 & 0.000000 \\
Model 2 & -72.267119 & 0.000000 \\
Model 3 & -68.972703 & 0.000000 \\
Model 4 & -71.595163 & 0.000000 \\
Model 5 & -75.631103 & 0.000000 \\
Model 6 & -69.794548 & 0.000000 \\
Model 7 & -72.959298 & 0.000000 \\
Model 8 & -62.863070 & 0.000000 \\
Model 9 & -73.555998 & 0.000000 \\
Model 10 & -64.343782 & 0.000000 \\
\midrule
\multicolumn{3}{c}{ROS} \\
\midrule
 & t-statistic & p-value \\
\midrule
Ensemble & -4.421883 & 0.000010 \\
Model 1 & -1.645748 & 0.099920 \\
Model 2 & -3.598124 & 0.000325 \\
Model 3 & -2.312506 & 0.020818 \\
Model 4 & -2.653743 & 0.008002 \\
Model 5 & -4.820706 & 0.000002 \\
Model 6 & -5.196932 & 0.000000 \\
Model 7 & -3.603603 & 0.000319 \\
Model 8 & -1.033712 & 0.301347 \\
Model 9 & -4.809240 & 0.000002 \\
Model 10 & 0.979096 & 0.327611 \\
\midrule
\multicolumn{3}{c}{RRS} \\
\midrule
 & t-statistic & p-value \\
\midrule
Ensemble & -40.463858 & 0.000000 \\
Model 1 & -9.543686 & 0.000000 \\
Model 2 & -6.262358 & 0.000000 \\
Model 3 & -8.260879 & 0.000000 \\
Model 4 & -9.029468 & 0.000000 \\
Model 5 & -8.938158 & 0.000000 \\
Model 6 & -9.841412 & 0.000000 \\
Model 7 & -5.519696 & 0.000000 \\
Model 8 & -9.235282 & 0.000000 \\
Model 9 & -12.332399 & 0.000000 \\
Model 10 & -5.896142 & 0.000000 \\
\bottomrule
\end{tabular}
\label{tab:t-test-gcamvsrandom2}
\end{table}

\end{document}